\documentclass{article}
\usepackage{ijcai22}

\usepackage{times}

\usepackage{soul}
\usepackage{url}
\usepackage[hidelinks]{hyperref}
\usepackage[utf8]{inputenc}
\usepackage[small]{caption}
\usepackage{graphicx}
\usepackage{amsmath}
\usepackage{amssymb}
\usepackage{amsthm}
\usepackage{booktabs}
\usepackage{xcolor}
\usepackage[ruled]{algorithm2e}
\usepackage{multirow}
\usepackage{threeparttable}
\usepackage{subfigure}
\usepackage{makecell}
\usepackage{xcolor,soul}
\usepackage{lipsum}
\usepackage{enumerate}
\usepackage{rotating}

\title{A Survey on Gradient Inversion: Attacks, Defenses and Future Directions}

\author{
Rui Zhang$^1$\and
Song Guo$^{1,2}$\footnote{Corresponding Authors}\and
Junxiao Wang$^{1*}$\and
Xin Xie$^1$\And
Dacheng Tao$^3$\\
\affiliations
$^1$ The Hong Kong Polytechnic University\\
$^2$ The Hong Kong Polytechnic University Shenzhen Research Institute\\
$^3$ JD Explore Academy, JD.com\\
\emails
csrzhang1@comp.polyu.edu.hk,
\{song.guo, junxiao.wang, xin-ryan.xie\}@polyu.edu.hk,
dacheng.tao@gmail.com
}

\begin{document}

\maketitle

\begin{abstract}
    Recent studies have shown that the training samples can be recovered from gradients, which are called \textit{\textbf{Grad}ient} \textit{\textbf{Inv}ersion} (\textbf{GradInv}) attacks. However, there remains a lack of extensive surveys covering recent advances and thorough analysis of this issue. In this paper, we present a comprehensive survey on GradInv, aiming to summarize the cutting-edge research and broaden the horizons for different domains. Firstly, we propose a taxonomy of GradInv attacks by characterizing existing attacks into two paradigms: \textit{iteration-} and \textit{recursion-}based attacks. In particular, we dig out some critical ingredients from the iteration-based attacks, including \textit{data initialization}, \textit{model training} and \textit{gradient matching}. Second, we summarize emerging defense strategies against GradInv attacks. We find these approaches focus on three perspectives covering \textit{data obscuration}, \textit{model improvement} and \textit{gradient protection}. Finally, we discuss some promising directions and open problems for further research.
\end{abstract}

\section{Introduction}

Distributed learning or federated learning \cite{mcmahan2017communication,bonawitz2019towards} has become a popular paradigm to achieve collaborative training and data privacy at the same time. In a centralized training process, the parameter server initially sends a global model to each participant. After training with local data, the participants are only required to share gradients for model updates. Then the server aggregates the gradients and transmits the updated model back to each user. However, recent studies have shown that gradient sharing is not as secure as it is supposed to be. We consider an \textit{honest-but-curious} attacker, who can be the centralized server or a neighbor in decentralized training. The attacker can observe gradients of a victim at time $t$, and he/she attempts to recover data $\mathbf{x}(t)$ or labels $\mathbf{y}(t)$ from gradients. In general, we name such attacks as \textit{{Grad}ient} \textit{{Inv}ersion} ({GradInv}) attacks.

A majority of GradInv attacks \cite{zhu2019deep,geiping2020inverting,yin2021see} purpose to minimize the distance between the generated gradients and ground-truth gradients. In order to generate dummy gradients, a pair of random data and labels are fed to the global model. Taking the distance between the gradients as error and the dummy inputs as parameters, the recovery process can be formulated as an iterative optimization problem. When the optimization procedure converges, the private data is supposed to be fully reconstructed. Moreover, some newly presented studies \cite{fan2020rethinking,zhu2021rgap,chen2021understanding} can also reconstruct the original data in a closed-form algorithm. The key insight is to exploit the implicit relationships among the input data, model parameters and gradients of each layer, and find the optimal solution with minimum error.

To prevent attackers from disclosing privacy from gradient sharing, some cryptography-based methods \cite{bonawitz2017practical,phong2018privacy}, differential privacy-based approaches \cite{sun2021soteria,wei2021gradient,nasr2021adversary} and pixel-based perturbations \cite{fan2019differential,huang2020instahide} have been proposed to enhance the security and privacy levels. In addition, such privacy leakage can be mitigated by increasing the local iterations or batch sizes \cite{wei2020framework,huang2021evaluating} during model training.

However, reviewing the existing surveys of privacy attacks and defenses in distributed learning \cite{bouacida2021vulnerabilities,jegorova2021survey,wainakh2021federated}, we found that there remains a lack of systematic and exhaustive summaries of recent GradInv research. Hence, in this paper, we present a comprehensive survey covering the attacks, defenses and future directions of GradInv. In a nutshell, this paper makes the following key contributions.

\begin{figure*}[htbp]
\centering
\subfigure[Workflow of iteration-based GradInv attacks]{
\centering
\includegraphics[height=5.075cm]{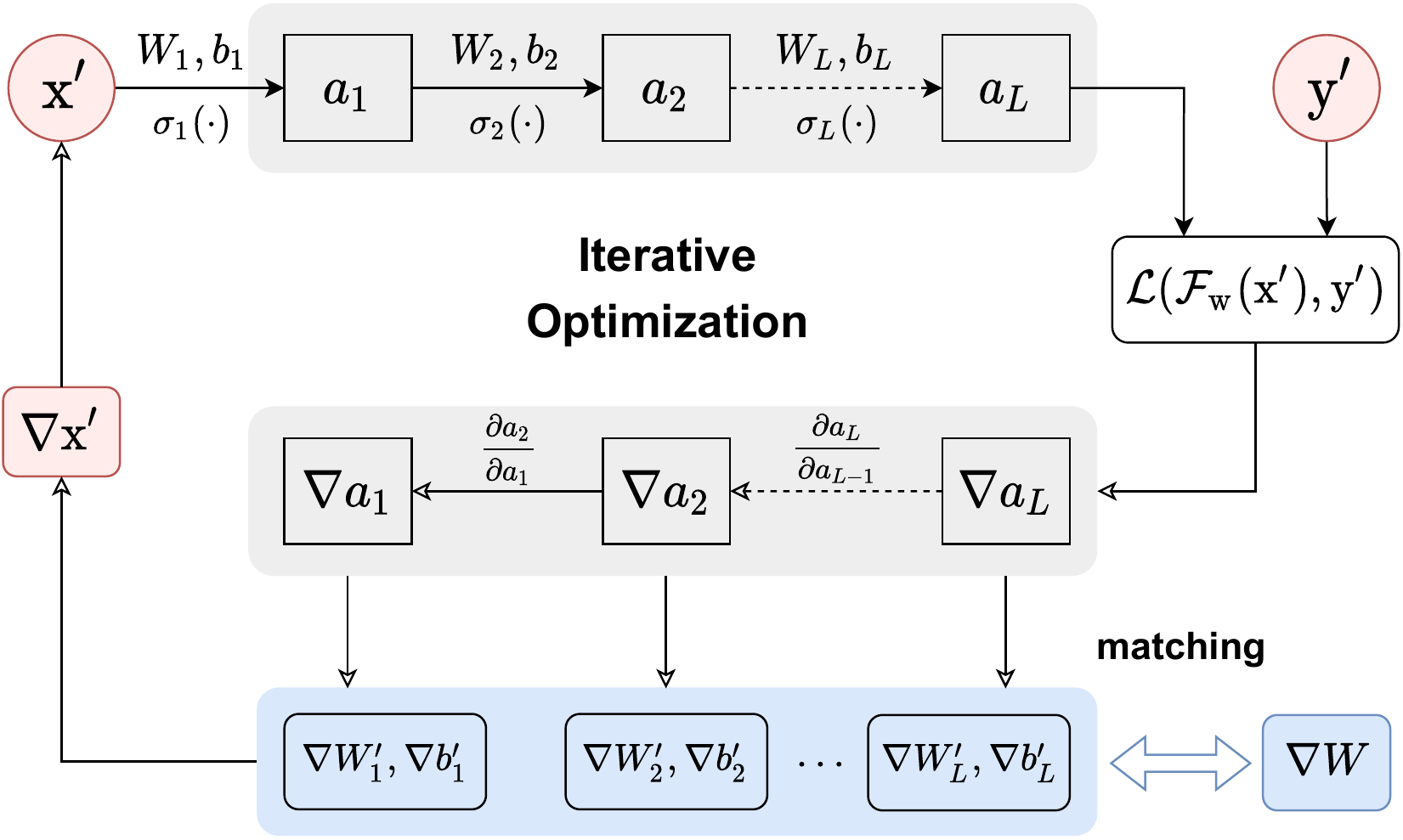}
\label{fig:iteration}}
\hfill
\subfigure[Workflow of recursion-based GradInv attacks]{
\centering
\includegraphics[height=5.075cm]{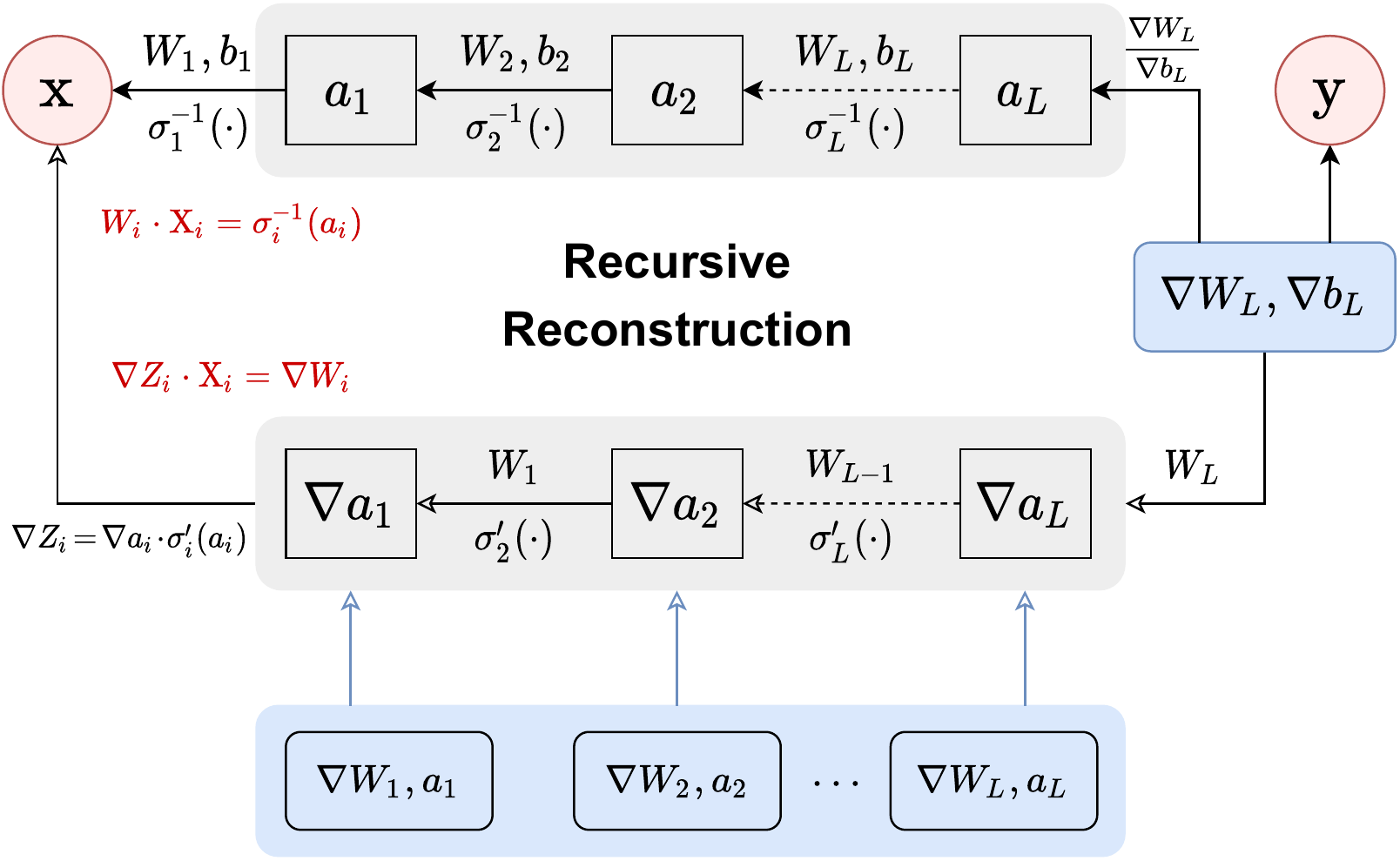}
\label{fig:recursion}}
\caption{Reconstruction workflow of two types of GradInv attacks. Both procedures include data, model and gradients, which are represented with different colors. (\textcolor[RGB]{184,84,80}{\textbf{Red}}: Reconstructed data, labels or gradients of the dummy inputs. \textcolor{gray}{\textbf{Gray}}: Global model shared in distributed learning, whose parameters and structure are known. \textcolor[RGB]{100,142,191}{\textbf{Blue}}: Ground-truth gradients and generated gradients for data recovery.)}
\label{fig:flow}
\end{figure*}

\begin{itemize}
    \item We first propose a novel taxonomy of GradInv attacks. Depending on the optimization objective, we categorize the existing studies into \textit{iteration-based} and \textit{recursion-based} attacks. The former aims at iteratively minimizing the distance between gradients, and the latter recursively recovers the optimal input layer by layer. Specially, we dig out some critical ingredients from the iteration-based attacks, including \textit{data initialization}, \textit{model training} and \textit{gradient matching}. We also analyze the impacts of these ingredients on the attacking effect.
    \item Then we summarize the emerging and representative defense strategies from the viewpoints of \textit{data obscuration}, \textit{model improvement} and \textit{gradient protection}. In particular, each participant can either obscure original images from a data source, improve the networks from security analysis, or protect the gradients before sharing.
    \item We finally conclude this survey and discuss some open problems and promising directions, which we hope can shed light on future research. Our proposed directions include the combination of two GradInv attacks, extension to Transformer architecture, and risk quantification between private data and shared gradients.
    \item To the best of our knowledge, this is the first comprehensive survey focusing on wide-ranging issues of GradInv.
\end{itemize}

\section{Gradient Inversion Attacks}
\label{sec:attack}

In this section, we provide a taxonomy of GradInv attacks by characterizing the existing works into two paradigms: \textit{iteration} and \textit{recursion}-based attacks. Moreover, we delve into the iteration-based attacks and divide the main components into \textit{data initialization}, \textit{model training} and \textit{gradient matching}. We first describe the workflow of each category and formulate the optimization objectives. Then we introduce the representative studies. A detailed comparison of differences and similarities among recent GradInv attacks is listed in Table~\ref{tb:attack}.

\subsection{Iteration-based Data Recovery}

In the iteration-based workflow, the attacker first generates a pair of random data $\mathbf{x}'$ and labels $\mathbf{y}'$, which are regarded as optimization parameters for data recovery. After forward and backward propagation, generated gradients of model weights $\nabla\mathrm{W}'$ can be obtained. Here, the bias terms are ignored since they can be integrated into the networks by adding a neuron with the input of 1. According to the distance between generated gradients and ground-truth gradients of a victim, another backward propagation is performed to calculate the gradients of dummy inputs, i.e., $\nabla\mathbf{x}'$ and $\nabla\mathbf{y}'$. From the perspective of reconstructed data, the reconstruction process includes once forward propagation, twice backward propagation and the update of dummy inputs, which is illustrated in Fig.~\ref{fig:iteration}.

The recovery of private data can be viewed as an iterative optimization process utilizing gradient descent. When the optimization converges, that is, the distance (e.g., $\ell_2$ norm) between gradients is close, and the original data is supposed to be fully recovered. The optimization problem is formulated in Eq.~(\ref{eq:iteration}), where $\mathbf{x}'^{*}, \mathbf{y}'^{*}$ are the optimized results.
\begin{equation}
\label{eq:iteration}
\begin{aligned} 
\mathbf{x}'^{*}, \mathbf{y}'^{*}
&=\underset{\mathbf{x}',~\mathbf{y}'}{\arg \min }\left\|\nabla\mathrm{W}'-\nabla\mathrm{W}\right\|^{2} \\
&=\underset{\mathbf{x}',~\mathbf{y}'}{\arg \min }\left\|\frac{\partial\mathcal{L}\left(\mathcal{F}\left(\mathbf{x}',\mathrm{W}\right), \mathbf{y}'\right)}{\partial\mathrm{W}}-\nabla\mathrm{W}\right\|^{2}
\end{aligned}
\end{equation}

\begin{table*}[ht]
\newcommand{\xiao}{\fontsize{9pt}{\baselineskip}\selectfont}
\renewcommand\arraystretch{1.3}
\centering
\begin{threeparttable}
\centering
\xiao
\begin{tabular}{ l | c  c | c  c | c  c | c }
\toprule
\multicolumn{1}{c|}{\multirow{2}{*}{\textbf{Publication}}} & \multicolumn{2}{c|}{\textbf{Data Initialization}} & \multicolumn{2}{c|}{\textbf{Model Training}} & \multicolumn{2}{c|}{\textbf{Grad Matching}} & \multicolumn{1}{c}{\multirow{2}{*}{\textbf{Additional}}} \\
 & \textit{Distribution} & \textit{Resolution} & \textit{Network} & \textit{Batch size} & \textit{Loss-fn} & \textit{Optimizer} &  \\
\midrule
\multicolumn{8}{c}{\textit{\textbf{GradInv attacks of iteration-based framework}}} \\
\midrule
    DLG \cite{zhu2019deep}  &  Gaussian  &  64$\times$64  &  LeNet  &  8  &  $\ell_2$ dist  &  L-BFGS  &  --  \\
    iDLG \cite{zhao2020idlg}  &  Uniform\tnote{$\mathbb{L}$}  &  32$\times$32  &  LeNet  &  1  &  $\ell_2$ dist  &  L-BFGS  &  --  \\
    CPL \cite{wei2020framework}  &  Geometric  &  128$\times$128  &  LeNet  &  8  &  $\ell_2$ dist  &  L-BFGS  &  $\mathcal{R}_{\text{y}}$ regularizer  \\
    InvGrad \cite{geiping2020inverting}  &  Gaussian\tnote{$\mathbb{L}$}  &  224$\times$224  &  ResNet\tnote{$\mathbb{T}$}  &  8 (100)  &  Cosine  &  Adam  &  $\mathcal{R}_{\text{TV}}$ regularizer  \\
    SAPAG \cite{wang2020sapag}  &  Constant  &  224$\times$224  &  ResNet\,\tnote{$\mathbb{T}$}  &  8  &  Gauss  &  AdamW  &  Gaussian kernel  \\
    GradInversion \cite{yin2021see}  &  Gaussian\tnote{$\mathbb{L}$}  &  224$\times$224  &  ResNet\tnote{$\mathbb{T}$}  &  48  &  $\ell_2$ dist  &  Adam  &  $\mathcal{R}_\text{fidel}$\ +\ $\mathcal{R}_\text{group}$  \\
    GradDisagg \cite{lam2021gradient}  &  Gaussian  &  32$\times$32  &  MLP  &  32 (128)  &  $\ell_2$ dist  &  L-BFGS  &  Participant info  \\
    GradAttack \cite{huang2021evaluating}  &  Gaussian\tnote{$\mathbb{L}$}  &  224$\times$224  & ResNet\tnote{$\mathbb{T}$}  &  128  &  Cosine  &  Adam  &  No BN + labels  \\
    Bayesian \cite{balunovic2022bayesian}  &  Gaussian  &  32$\times$32  &  ConvNet\tnote{$\mathbb{T}$}  &  1 (32)  &  Cosine  &  Adam  &  Known $p(g|x)$  \\
    CAFE \cite{jin2021catastrophic}  &  Uniform  &  32$\times$32  &  Loop-Net  &  100  &  $\ell_2$ dist  &  SGD  &  In Vertical-FL  \\
    GIAS \cite{jeon2021gradient}  &  Latent  &  64$\times$64  &  ResNet\tnote{$\mathbb{T}$}  &  4  &  Cosine  &  Adam  &  GAN-based  \\
\midrule
\multicolumn{8}{c}{\textit{\textbf{GradInv attacks of recursion-based framework}}} \\
\midrule
    PPDL-AHE \cite{phong2018privacy}  &  N/A  &  20$\times$20  &  MLP  &  1  &  \multicolumn{2}{c|}{Gradient division}  &  --  \\
    PPDL-SPN \cite{fan2020rethinking}  &  N/A  &  32$\times$32  &  ConvNet  &  8  &  \multicolumn{2}{c|}{Linear solving}  &  Noise analysis  \\
    R-GAP \cite{zhu2021rgap}  &  N/A  &  32$\times$32  &  ConvNet  &  1  &  \multicolumn{2}{c|}{Inverse matrix}  &  Rank analysis  \\
    COPA \cite{chen2021understanding}  &  N/A  &  32$\times$32  &  ConvNet  &  1  &  \multicolumn{2}{c|}{Least-squares}  &  Pull-back const  \\
\bottomrule
\end{tabular}
\begin{tablenotes}
\scriptsize
    \item[$\mathbb{L}$] The labels can be directly identified or extracted from shared gradients.
    \item[$\mathbb{T}$] The results of data recovery are compared in different model training states.
\end{tablenotes}
\end{threeparttable}
\caption{Summary and classification of existing GradInv attacks.}
\label{tb:attack}
\end{table*}

\subsubsection{Initialization of Input Data and Labels}
\label{sec:init}

To produce dummy gradients and perform gradient matching, the attacker first needs to generate random data and labels. Initialization involves the selection of distribution, the size of original data, and the asynchronous recovery of labels.

\paragraph{Distribution.}
Random Gaussian noise is most frequently used for data initialization in the majority of GradInv attacks \cite{zhu2019deep,geiping2020inverting,yin2021see}. In addition, constant values \cite{wang2020sapag} or random noise sampled from Uniform distribution \cite{jin2021catastrophic} are also presented for data initialization. However, some experiment results demonstrate that GradInv attacks often fail to converge due to bad initialization \cite{geiping2020inverting}. Hence, \cite{wei2020framework} prove that the convergence of a reconstruction process can be guaranteed by Lipschitz continuity:
\begin{equation}
\label{eq:init}
f(\mathbf{x}'^{\mathrm{T}})-f(\mathbf{x}'^{*}) \leq \frac{2L\left\|\mathbf{x}'^{0} - \mathbf{x}'^{*}\right\| ^{2}}{\mathrm{T}}
\end{equation}
where $L$ is the Lipschitz constant, $\mathrm{T}$ denotes the condition of termination, and $\mathbf{x}'^{0}$, $\mathbf{x}'^{\mathrm{T}}$, $\mathbf{x}'^{*}$ represent the initializing data, terminated results and optimal recovery, respectively. Based on Eq.~(\ref{eq:init}), the authors propose patterned randomization and theoretical optimal initialization methods, which are more efficient and stable than Gaussian or Uniform noise.

\paragraph{Image Resolution.}
Without loss of generality, we consider the scenarios of image classification tasks. Actually, the resolution of raw images is an important factor that affects both initialization and the difficulty of recovery. The more pixels an image has, the more variables need to be optimized. Thus, it is more challenging to deploy attacks on complex datasets, such as ImageNet \cite{deng2009imagenet}. In contrast, relatively good results can be achieved on the low-resolution black-and-white images (e.g., MNIST \cite{lecun1998mnist}, Fashion-MNIST \cite{xiao2017fashion}), even under mini-batch training. Till now, 224$\times$224 pixels is the largest resolution for recovery.

\paragraph{Label Extraction.}
In a general process for iteration-based GradInv attacks, dummy data and labels are simultaneously updated. However, if the ground-truth labels can be extracted in advance, data recovery will be accelerated and the computational complexity will also be reduced. \cite{zhao2020idlg} first find out that the ground-truth label in a classification task can be directly revealed. For a mini-batch recovery, \cite{sun2021soteria,yin2021see} propose that labels of different classes can be identified from each other, which assumes that there are non-repeating labels in the mini-batch training data. All of the above approaches extract the labels from the fully connected layer of gradients. Moreover, \cite{wainakh2021user} exploit both the angle and magnitude of gradients to identify the labels. The angle shows whether a label is contained in the batch, while the magnitude indicates the number of duplicate labels. \cite{dang2021revealing} present a more powerful label leakage attack, which can be applied to both image classification and speech recognition tasks. The labels can still be revealed after multiple local iterations.

\subsubsection{Model Training for Gradient Generation}
\label{sec:model}

In order to obtain the generated gradients, the attacker needs to feed the initialized dummy data and labels into the model. Based on the error between the model outputs and the labels, the gradients of weights can be calculated through backpropagation. In the setting of distributed learning, the global model can be viewed as a white box, which means the model structure and weights are known. However, the depth of training network structures and the batch size of training data can implicitly affect the results of data recovery.

\paragraph{Network Model.}
Convolutional neural networks (CNN) or multilayer perceptron (MLP) are generally adopted as training networks for computer vision. Intuitively, the deeper the network is, the more parameters it contains. This raises two serious issues. First, the computational complexity is greatly increased, which makes it difficult or impossible for the optimization process to converge. Second, even if the procedure converges, there may exist multiple locally optimal solutions, resulting in a significant difference in the ground-truth value.
\newline
So far, \cite{geiping2020inverting} are able to recover the raw data on ResNet-152 \cite{he2016deep}, although only a few images can be recognized. \cite{yin2021see} achieve data recovery of high-resolution images on ResNet-50, and display relatively better results. \cite{jin2021catastrophic,balunovic2022bayesian} also investigate the relationship between the convergence states of model training and the errors of reconstruction.

\paragraph{Batch Training.}
For regular training, learning with mini-batch data can decrease the number of iterations and reduce the fluctuation of accumulated errors. However, this greatly increases the difficulty of GradInv attacks. Given an observation of gradients, the problem of recovering original data is equivalent to the decomposition of averaged summation. Using the vanilla optimization approach in Eq.~(\ref{eq:iteration}), \cite{zhu2019deep} can only perform data recovery for a maximum batch size of 8. With the assistance of various regularization terms, \cite{yin2021see,huang2021evaluating} all can recover the original private data with a batch size of over 30.

\subsubsection{Data Update through Gradient Matching}

The process of gradient matching measures the difference between generated gradients and ground-truth gradients, and then calculates the update of dummy inputs. Essentially, the procedure for data recovery can be analogized to supervised learning, which means the ground-truth gradients are similar to a high-dimensional ``label", while the dummy data and labels are the parameters to be learned in the optimization process. There are several important parts for gradient matching, the first is to obtain the gradients of a victim, and the next is proposing an effective method to minimize the distance.

\paragraph{Disaggregation.}
Considering the secure aggregation rules \cite{bonawitz2017practical,fereidooni2021safelearn}, the attacker can only observe a summation of the gradients. To perform gradient matching, it is necessary to decompose the individual gradients from the aggregation. Similar to the decomposition problem of mini-batch recovery, this is also a challenging problem. \cite{lam2021gradient} first focus on this issue and formulate it as a matrix factorization problem. By leveraging the participant information acquired from device analytics, additional constraints contribute to the solution. If $G_\text{agg}$ represents the aggregated gradients, the factorization problem can be solved by finding the participant vector $p_k$:
\begin{equation}
\begin{aligned}
\text{Find} \ \ p_{k} \ \ \text{s.t.} \ \, & \operatorname{Null}(G_\text{agg}^{\mathsf{T}}) p_{k}=0 \quad \\
& \, C_{k} p_{k}-c_{k}=0 \\
& \,\, p_{k} \in\{0,1\}^{n}
\end{aligned}
\end{equation}
where $\operatorname{Null}(\cdot)$ calculates the kernel of a matrix, and $C_{k}$ specifies the participated rounds of client $k$.

\paragraph{Loss Function.}
Generally, an attacker is assumed to have direct access to the gradients of a victim. Thus, these studies focus on improving their algorithms to decrease the difference between gradients, and recover more realistic data. The enhancement consists of optimization metrics and regularization terms. To measure the distance between generated gradients and ground-truth gradients, Euclidean distance (i.e., $\ell_2$ norm) is a frequently used loss function \cite{zhu2019deep,wei2020framework,lam2021gradient}. However, \cite{wang2020sapag} discover that gradients with large values dominate data recovery at the early stages. They hence propose a weighted Gaussian kernel as the distance metric. Furthermore, \cite{geiping2020inverting} observe that the direction of gradients plays a more important role than magnitude, and substitute the $\ell_2$ cost function with cosine similarity:
\begin{equation}
\arg \min_{\mathbf{x}'\in[0,1]^{n}} 1-\frac{\left\langle\nabla\mathrm{W}',\nabla\mathrm{W}\right\rangle}
{\left\|\nabla\mathrm{W}'\right\| \left\|\nabla\mathrm{W}\right\|}
\end{equation}


\paragraph{Regularization.}
To recover more realistic data from the batch, some auxiliary regularization terms are inserted into the cost functions. These regularizers can be divided into two categories, one that constrains the fidelity of images, and the other that revises the position of the main object. Fidelity regularization \cite{geiping2020inverting,yin2021see} steers the reconstructed data from impractical images, including total variation norm (TV), $\ell_2$ norm and batch normalization (BN) \cite{nguyen2015deep,mahendran2015understanding}. Group consistency regularization is presented by \cite{yin2021see}, which jointly considers multiple random seeds for initialization, and calculates the averaged data $\mathbb{E}(\mathbf{x}'_{g \in G})$ as reference. Any candidate whose recovery deviates from the ``consensus'' image of the group will be penalized.
\begin{equation}
\begin{aligned}
& \mathcal{R}_{\text{fidel}} ({\mathbf{x}'})=\alpha_{\mathrm{tv}} \mathcal{R}_{\mathrm{TV}}({\mathbf{x}'})+\alpha_{\ell_{2}} \mathcal{R}_{\ell_{2}}({\mathbf{x}'})+\alpha_{\mathrm{bn}} \mathcal{R}_{\mathrm{BN}}({\mathbf{x}'}) \\
& \mathcal{R}_{\text{group}}({\mathbf{x}'}, \mathbf{x}'_{g \in G})=\alpha_{\text{group }}\!\!\left\|\mathbf{x}'-\mathbb{E}(\mathbf{x}'_{g \in G})\right\|^{2}
\end{aligned}
\end{equation}

Applying $\mathcal{R}_{\text{fidel}}$ and $\mathcal{R}_{\text{group}}$ regularizers can indeed support reconstructing more realistic image data on complex datasets (e.g., ImageNet) under mini-batch training.

\begin{table*}[htbp]
\newcommand{\zhong}{\fontsize{9.35pt}{\baselineskip}\selectfont}
\renewcommand\arraystretch{1.425}
\centering
\zhong
\begin{tabular}{ p{2.6cm}<{\centering} p{2.2cm}<{\centering} l l }
\specialrule{0.825pt}{0pt}{2.5pt}
\textbf{Category} & \textbf{Method} & \multicolumn{1}{c}{\textbf{Publication\quad}} & \multicolumn{1}{c}{\textbf{Key Contribution\quad}} \\
\specialrule{0.55pt}{2.5pt}{0.2pt}
\multicolumn{1}{c}{\multirow{3}{*}{\textbf{\textit{Original Data}}}}
    &  MixUp  &  \cite{zhang2018mixup}  &  Data enhancement by linearly combining the inputs \\ \cline{2-4}
    &  InstaHide  &  \cite{huang2020instahide}  &  Encrypt the MixUp data with one-time secret keys \\ \cline{2-4}
    &  Pixelization  &  \cite{fan2018image,fan2019differential}  &  Perturb the raw data with pixelization-based method \\
\specialrule{0.4pt}{0.2pt}{0.2pt}
\multicolumn{1}{c}{\multirow{3}{*}{\textbf{\textit{Training Model}}}}
    &  \multirow{1}{*}{Dropout}  &  \cite{zheng2021dropout}  &  Add an additional dropout layer before the classifier \\ \cline{2-4}
    &  \multirow{1}{*}{Local iters}  &  \cite{wei2020framework}  &  Share gradients after multiple local training iterations \\ \cline{2-4}
    &  \multirow{1}{*}{Architecture}  &  \cite{zhu2021rgap}~~~  &  Reduce the number of convolutional kernels properly \\
\specialrule{0.4pt}{0.2pt}{0.2pt}
\multicolumn{1}{c}{\multirow{6}{*}{\textbf{\textit{Shared Gradients}}}}
    &  \multirow{2}{*}{Aggregation}  &  \cite{zhang2020batchcrypt}  &  Apply Homomorphic Encryption to protect gradients \\ \cline{3-4}
    &  &  \cite{lia2020privacy}  &  Utilize Secure Multi-Party Computation to aggregate\qquad \\ \cline{2-4}
    &  \multirow{2}{*}{Perturbation}  &  \cite{sun2021soteria}  &  Perturb data representation layer and maintain utility \\ \cline{3-4}
    &  &  \cite{wei2021gradient}  &  Add adaptive noise with differential privacy guarantee~~ \\ \cline{2-4}
    &  \multirow{2}{*}{Compression}  &  \cite{vogels2019powersgd}  &  Compress the smaller values in gradients to zero \\ \cline{3-4}
    &  &  \cite{karimireddy2019error}  &  Transmit the sign of gradients for model updates \\
\specialrule{0.825pt}{0.2pt}{0pt}
\end{tabular}
\caption{Summary and classification of existing GradInv defense strategies.}
\label{tb:defense}
\end{table*}

\subsection{Recursion-based Data Recovery}

In recursion-based data recovery, the attacker can recursively calculate the input of each layer by finding the optimal solution with minimized error. \cite{phong2018privacy} first discover that the input of a perceptron can be directly recovered from: $x_k=\nabla\mathrm{W}_k/\nabla b$. This conclusion is later generalized to fully connected (FC) layers or MLP, as long as bias terms exist. In accordance with this idea, \cite{fan2020rethinking} convert a convolutional layer into a FC layer by stacking the filters, and then utilize the above relation. However, they neglect the feature of reused weights in the convolutional layers, which is totally different from that of FC layers. In order to recover the image data of the first convolutional layer, \cite{zhu2021rgap} combine the forward and backward propagation, and formulate the problem as solving a system of linear equations. The essence of such data leakage is that: the feature map and the gradient of kernels in the first convolutional layer have direct involvement with the original data. Extending to the $i$th layer, it is possible to recover the input $\mathbf{x}_i$ by solving:
\begin{equation}
\label{eq:recursion}
\left\{ \ 
\begin{aligned}
\vspace{2pt}
\mathrm{W}_i \cdot \mathbf{x}_i &= Z_i \\
\nabla Z_i \cdot \mathbf{x}_i &= \nabla\mathrm{W}_i
\end{aligned}
\right.
\end{equation}
where $Z_i$, $\nabla Z_i$ represent the feature map and its gradients. Denoting the neuron outputs and activation function of each layer as $a_i$ and $\sigma_i(\cdot)$, then we have the relationships of $Z_i=\sigma_i^{-1}(a_i)$ and $\nabla Z_i=\nabla a_i\cdot \sigma'_i(a_i)$. Hence, it is possible to recover the original data by recursively starting from the FC layer to the convolutional layer. \cite{chen2021understanding} also propose a generic framework by combining multiple optimization problems under different situations. The workflow of the recursion-based framework is depicted in Fig.~\ref{fig:recursion}


\newpage

Compared to the iteration-based approach, these recursion-based attacks have the following characteristics.

\paragraph{Non-initialization.}
Different from iterative optimization, a recursion-based attack can directly recover the original image without initialization or generating dummy inputs. Since the time of recovery is proportional to the square of the number of pixels, the current image resolution that can be recovered does not exceed 32$\times$32. Furthermore, the extraction of labels refers to the above-mentioned works in Section~\ref{sec:init}.

\paragraph{Model Structure.}
The networks for experiments include only convolutional and fully connected layers. The pooling layers or shortcut connections in ResNet are not considered, which would cause an accumulation of errors layer by layer, and make the reconstructed data extremely biased. However, these attacks can not deal with mini-batch training, which is a major difference compared to iteration-based attacks.

\paragraph{Linear Solving.}
As formulated in Eq.~(\ref{eq:recursion}), the original data can be recursively solved by constructing linear equations. The feature maps and their gradients can be derived from the model weights and corresponding gradients. It is worth mentioning that such an attack depends on the integrity of gradients. Once the gradients are perturbed, the noisy solution will make the restored results completely unrecognizable.


\section{Gradient Inversion Defenses}

In this section, we summarize the emerging defense strategies from perspectives of \textit{data obscuration}, \textit{model improvement}, and \textit{gradient protection}. A client can either obscure the private images from the data source, enhance the structure of network models, or protect the gradients before sharing. The main contributions of these defenses are described in Table~\ref{tb:defense}.

\subsection{Obscuration of Original Data}

Since the target of GradInv attacks is to recover the training data of a victim, an ideal defense strategy is to directly protect the raw data before training. We expect that the private input is difficult to be reconstructed while the model utility does not degrade too much. \cite{zhang2018mixup} propose the method of MixUp for data augmentation, where virtual training samples are generated by linearly combining a pair of data and labels. These generated examples can not only improve the accuracy of the training model, but also ``aggregate'' the original data. Based on MixUp, \cite{huang2020instahide} introduce the idea of cryptography to protect the data using one-time private keys. In particular, a portion of images from both private and public datasets are randomly selected for combination, and then the pixels are flipped according to the keys. This lightweight approach prevents an attacker from recovering the training data and ensures the usability of the data. \cite{pang2019mixup} also mixup the input with other clean samples to improve the adversarial robustness of training models. In addition, \cite{fan2019differential} protect the images with pixelization and Gaussian blur approaches, which can be used not only for distributed training, but also for data publication, such as crowd-sourcing.


\subsection{Improvement of Training Model}

As for training models, in addition to increasing the depth of the neural network or training with mini-batch (Section~\ref{sec:model}), we introduce some newly presented but effective approaches. In a general GradInv attack, it is assumed that the gradients are sent after one round of local training. \cite{wei2020framework} propose to schedule and control the number of locally training iterations before gradient sharing, which makes it more difficult to reconstruct the private data. Experiments demonstrate that the success rates of data recovery have dropped by more than 60\% when performing 10 local iterations. Modifications to the network structure can also defend against the GradInv attacks. \cite{zheng2021dropout} propose to add a simple dropout layer between the encoder and the classifier to solve the problem of overfitting. During the training process, there may be certain neurons with larger activation values, indicating the features of training data are overly memorized. If a proportion of neurons are randomly pruned, then it is possible to mitigate the privacy inference attacks. Considering that different network models may have different risks of privacy leakage, \cite{zhu2021rgap} present a rank-based security analysis. Such method indicates that the more filters in a convolutional layer, the better the data recovery will be. Similarly, \cite{geiping2020inverting} have also mentioned that it is impossible to recover the original data from gradients, if the dimensionality of model parameters is lower than that of input data. This conclusion inspires us to appropriately reduce the number of parameters while ensuring the performance of the model.

\subsection{Protection from Gradient Sharing}

In distributed learning, since model updates are performed on the basis of gradient exchanging, the straightforward privacy-preserving approach is to protect the gradients. Summarizing the existing studies, we divide them into: {aggregation}, {perturbation} and {compression}-based defense strategies.

Cryptography-based methods can generally guarantee the security and privacy of individual gradients without compromising their utility. \cite{bonawitz2017practical,lia2020privacy} use secure multi-party computation (MPC) to compute the summation result of model updates. \cite{phong2018privacy,kim2018efficient,zhang2020batchcrypt} implement Homomorphic Encryption (HE) to carry out operations on the ciphertext space for gradient aggregation. Even if an attacker steals the information through man-in-the-middle (MITM) attacks, he/she cannot decrypt it to obtain the ground-truth gradients. However, these approaches not only require modifications to the training architecture, but also exponentially increase the computation time, bandwidth and data storage. 

Gradient perturbation is another frequently used approach for privacy protection. \cite{truex2020ldp,wei2020federated,yang2021h} add Gaussian noise to the exchanged gradients with the guarantee of differential privacy (DP). \cite{he2021tighter} essentially reveal how iterative training impacts privacy, and establishes the relationship between generalization and privacy-preserving. \cite{sun2021soteria} find the key to GradInv attacks lies in the data representation layer, and only perturb the gradient values in this layer. \cite{wei2021gradient} propose a method with dynamically adjustable noise which can achieve high resilience against GradInv attacks. Except for injecting noise, \cite{zhu2019deep} discover that some compression methods, originally used to reduce communication overhead, can also be used to prevent data recovery. \cite{vogels2019powersgd} propose to prune the smaller values to zero by a certain percentage, and \cite{karimireddy2019error} only transmit the sign of gradients. These methods can resist attacks to a certain extent while maintaining performance.

\section{Conclusion and Future Directions}

In this paper, we present recent advances in \textit{Gradient Inversion} (GradInv) covering the offensive and defensive methods. To the best of our knowledge, for the first time, we review the GradInv attacks with a novel taxonomy and describe the main procedure for deploying an attack in distributed learning. We also reclassify the representative defense strategies, which can be used to protect from data recovery. After summarizing the existing research on GradInv, we find that there are still some unresolved problems worthy of in-depth study. Finally, we discuss a few promising directions.

\subsubsection{Combination of GradInv Attacks}

We have introduced the characteristics of iteration-based and recursion-based attacks. The former can recover complex inputs from mini-batch data, while the latter enables one-shot and label-free reconstructions. \cite{qian2021minimal} present a two-step algorithm including both GradInv attacks, but only for a single convolutional layer without batch data. However, very few works have considered the combination of two kinds of attacks. If the constraint equations provided by recursion-based methods can be integrated into the iterative optimization procedure, we are able to improve the recovered results. Thus, it is necessary to propose a more powerful and efficient GradInv attack, which can speed up the recovery process or achieve a batch recovery on higher resolution datasets.

\subsubsection{Extension to Transformers}

Transformer \cite{vaswani2017attention,dosovitskiy2021an} is a recent popular network model that can be applied to both natural language processing and computer vision. Although GradInv attacks can be implemented on deep networks, such as ResNet, it is critical to extend from CNNs to Transformers. Since Transformer can handle more types of data, the attacks therein will cause more serious impacts. \cite{deng2021tag} first attempts to deploy GradInv attacks on Transformer-based language models. \cite{lu2021april} propose to recover input of self-attention modules in Vision Transformer theoretically and empirically. Although each of these works has some limitations, it remains crucial and urgent to focus on the security and privacy issues in Transformers.



\subsubsection{Quantification of Gradient Sharing}

The defenses we have discussed are all active interventions to the training process. However, we can also indirectly analyze whether the gradients have a serious risk of privacy leakage. Since each user owns his/her training data and gradients, the extent of leakage can be quantified if any correlation between them is found. \cite{liu2021quantitative} utilize mutual information to formulate this problem, and demonstrate the risk of gradient leakage is related to the model status and data distribution. \cite{mo2021layer} propose \textit{$\mathcal{V}$-information} and \textit{sensitivity} metrics to quantify the layer-wise latent privacy leakages. This indicates that research in other domains can be transferred to study GradInv. We hope this survey can shed a light on future research and inspire more progress in different domains.

\vspace{6pt}
\section*{Acknowledgments}

This research was supported by fundings from the Key-Area Research and Development Program of Guangdong Province (No. 2021B0101400003), Hong Kong RGC Research Impact Fund (No. R5060-19), General Research Fund (No. 152221/19E, 152203/20E, and 152244/21E), the National Natural Science Foundation of China (61872310), and Shenzhen Science and Technology Innovation Commission (JCYJ20200109142008673).

\newpage

\bibliographystyle{named}
\bibliography{ijcai22}

\end{document}